\documentclass{article}
\usepackage{ijcai21}

\usepackage{times}
\usepackage{soul}
\usepackage{url}
\usepackage[hidelinks]{hyperref}
\usepackage[utf8]{inputenc}
\usepackage[small]{caption}
\usepackage{graphicx}
\usepackage{amsmath}
\usepackage{amsthm}
\usepackage{amsfonts,amssymb}
\usepackage{booktabs}
\usepackage{algorithm}
\usepackage{algorithmic}

\usepackage{graphicx}
\usepackage{xcolor}
\usepackage{multirow}
\usepackage{amsmath}

\title{
\vspace*{-0.5in}
{{\small \hfill IJCAI'2021}\\
\vspace*{.25in}} 
Relational Gating for ``What If'' Reasoning}

\author{
Chen Zheng
\And
Parisa Kordjamshidi\\
\affiliations
Michigan State University\\
\emails
\{zhengc12, kordjams\}@msu.edu
}

\begin{document}

\maketitle

\begin{abstract}
This paper addresses the challenge of learning to do procedural reasoning over text to answer "What if..." questions. We propose a novel relational gating network that learns to filter the key entities and relationships and learns contextual and cross representations of both procedure and question for finding the answer. Our relational gating network contains an entity gating module, relation gating module, and contextual interaction module. These modules help in solving the "What if..." reasoning problem. We show that modeling pairwise relationships helps to capture higher-order relations and find the line of reasoning for causes and effects in the procedural descriptions. Our proposed approach achieves the state-of-the-art results on the WIQA dataset.

\end{abstract}

\section{Introduction}
\label{sec:intro}

\begin{figure}
\centering
\includegraphics[width=0.4\textwidth,height=180pt]{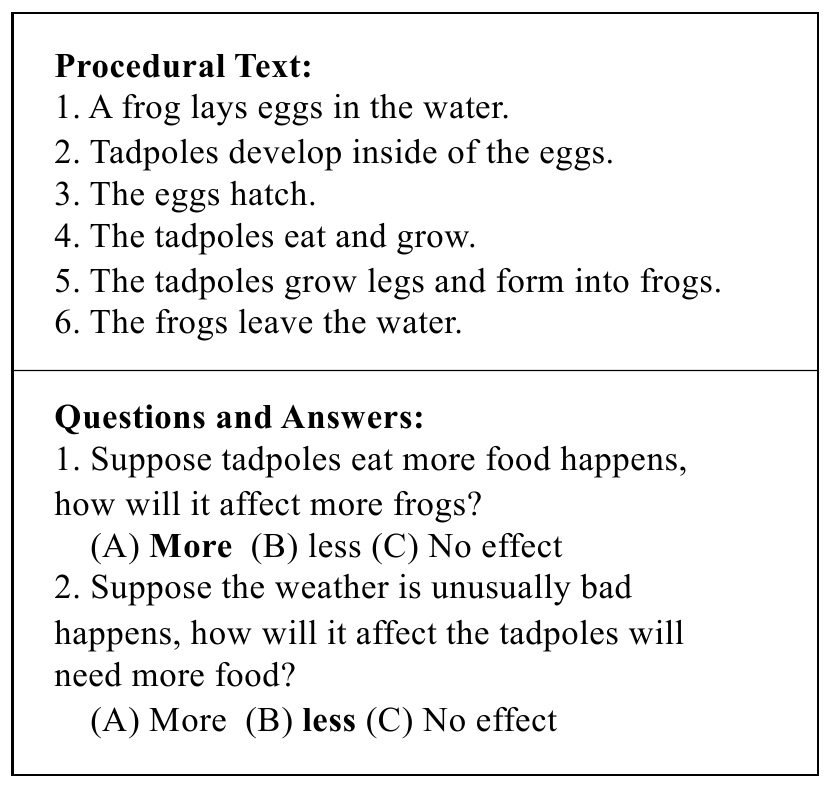}
\caption{WIQA task contains procedural paragraphs, and a large collection of ``what-if'' questions. The bold font candidate answers are the gold answers.} \label{fig:wiqa_example}
\end{figure}

The recent research on reasoning over procedural text has achieved promising results~\cite{Rajpurkar2016SQuAD10,Rajpurkar2018KnowWY,Henaff2017TrackingTW,Dalvi2018TrackingSC,Tandon2018ReasoningAA}.
Specific to this problem, a WIQA dataset and task~\cite{Tandon2019WIQAAD} was proposed as a benchmark for the evaluation of reasoning capabilities of learning models on procedural text by introducing 
``what \dots if'' questions.
The ``what \dots if'' reasoning problem is a procedural text QA that relates to reading comprehension, multi-hop reasoning, and commonsense reasoning which makes the task rich in containing various challenging linguistic and semantic phenomena.
Moreover, the ``what \dots if'' reasoning is built based on linguistic perturbations and generating possible cause-effect relationships expressed in the context of a paragraph. 
Its goal is to predict what would happen if a process was perturbed in some way. It requires understanding and tracing the changes in events and entities through a paragraph. Figure~\ref{fig:wiqa_example} shows some examples of WIQA task. 
There are three types of questions in the dataset, including in-paragraph where the answer to the question is in the procedure itself, out-of-paragraph where the answer does not exist in the text and needs external knowledge, and irrelevant (no effect) changes~\cite{Tandon2019WIQAAD}.
Given a procedural text, we answer questions with the ``what \dots if'' style.

There are several challenges in the ``what...if'' reasoning over procedural text. The first challenge is reasoning over the qualitative comparison expressions for describing the changes in the procedure that can make a positive or negative change. For example, reasoning over comparative words such as (larger, smaller), (more, less), (higher, lower). This task requires the ability to extract the important entities through the procedural text and understand their influences. Several previous works primarily use formal models for implicit qualitative comparisons~\cite{Parikh2016ADA}.
\citeauthor{Tandon2019WIQAAD} uses BERT to predict answers by implicit representations. However, they ignore explicit reasoning over qualitative comparisons between entities and the way they affect each other.

The second challenge is reasoning over relations between pairs of entities, we call it relational reasoning. Although recent pre-trained language models (LM) achieve promising performance on QA, there is still a gap between LM and human performance due to the lack of relational reasoning over entities~\cite{feng-etal-2020-scalable}.
For example, given the question ``suppose more animals that hunt frogs happens, how will it affect more tadpoles loses'', the LM is difficult to consider the relation ``hunt'' between the entity pair (``animals'', ``frogs'').
\citeauthor{asai2020logic} uses a Transformer model with regularization to produce consistent answers. The model obtains a good result with augmented data following logical constraints. However, these constraints ignore the importance of relational reasoning, and can not capture the higher-order chain of reasoning based on pairwise relations. 

The third challenge is the multi-hop reasoning. In-para and out-of-para question categories need multiple hops of reasoning to answer the questions. Like~\citeauthor{Tandon2019WIQAAD} said, given the question, indirect effects (2 hops or 3 hops) are much harder to answer than direct effects (1-hop).  For example, predicting ``cloudy day'' results in the direct effect in ``less sunshine'' is less challenging than the indirect effect in ``less photosynthesis''. 
Recently, \citeauthor{madaan2020eigen} presents EIGEN to leverage pre-trained language models to generate the reasoning chain.

The fourth challenge is the lexical variability in expressing the same concept, which makes entity alignment hard. For example, the same entities and events referred to by different terms, like (insect, bee), (become, form). Entity alignment requires the alignment between question and paragraph entities, and the alignment between the entities appearing in the different paragraphs themselves.
Unfortunately, all current works ignore the importance of entity alignment for tracing the entities and finding the relation between different entities in the question and paragraph.

Therefore, we propose a novel end-to-end Relational Gating Network (RGN) for procedural text reasoning. The RGN framework answers the procedural text question and solves challenges of qualitative comparison, relational reasoning, multi-hop reasoning, and entity alignment.
RGN jointly learns to extract the key entities through our entity gating mechanism, finds the line of reasoning and relations between the key entities through relation gating mechanism, and captures the entity alignment through contextual entity interaction. The main motivation of the two gating mechanisms is to learn the line of reasoning and to address the multi-hop reasoning challenge. This distinguishes the RGN model from the existing works.
Concretely, we build an \textit{entity gating} module to extract and filter the key entities in the question and context and highlight the entities that are compared qualitatively.
Furthermore, we design a \textit{relation gating} module with an alignment of crucial entities to capture the higher-order chain of reasoning based on pairwise relations. This technique helps relational reasoning and consequently multi-hop reasoning in the procedure. 
Moreover, we propose an efficient module, called \textit{contextual interaction module}, to incorporate cross information from Question and Content interactions during training in an efficient way to help entities alignments. 

The contributions of this work are as follows:
1) We propose a Relational Gating Network (RGN) that captures the most important entities and relationships involved in qualitative comparison, causal reasoning and multi-hop reasoning.
2) We propose a contextual interaction module to effectively and efficiently align the question and paragraph entities.
Moreover, the contextual interaction module captures the interacting entities and the change statements and helps to understand the qualitative comparisons. 
3) We evaluate the methods and analyze the results on ``what...if'' question answering using the WIQA dataset. We improve state-of-the-art on WIQA published results. We show the significance of the entity gating module and relation gating module on procedural reasoning over text.

\section{Related Work}
In recent years, large-scale Language Models~\cite{Devlin2019BERTPO,Liu2019RoBERTaAR} made a huge progress in solving various QA tasks on popular benchmarks such as SQuAD V1~\cite{Rajpurkar2016SQuAD10} and SQuAD V2~\cite{Rajpurkar2018KnowWY}.
However, in those benchmarks, there is no need for multi-hop reasoning and also the given text is sufficient to predict the answer. Therefore, the developed QA models fail to answer the questions that need procedural reasoning and understanding causes and effects~\cite{Dua2019DROPAR}. 
Therefore, several new QA  benchmarks created ~\cite{Dalvi2018TrackingSC,dalvi-etal-2019-everything} and aim to track the explicit entity states and find explainable answers given a text.
\citeauthor{Tandon2019WIQAAD} proposed the WIQA task that aims to solve ``what...if'' procedural reasoning and understand the effects of qualitative reasoning. 

Several previous works achieved impressive performance on procedural text reasoning~\cite{madaan2020eigen,huang2020rem,asai2020logic,Tandon2019WIQAAD}. Rem-Net~\cite{huang2020rem} uses a recursive memory network to find the answer. \citeauthor{rajagopal-etal-2020-ask} constructs the explanations based on the effects of perturbations in procedural text.
However, these models ignore the importance of capturing the higher-order chain of reasoning based on pairwise relations. However, on a different thread of work on visual question answering, \citeauthor{zheng-etal-2020-cross} considers the relevance beyond entities in multiple modalities and proposes a model to obtain the higher-order relevance of pairwise relations and shows its effectiveness.
Graph-based models~\cite{Velickovic2018GraphAN,zheng-kordjamshidi-2020-srlgrn} help in using the relational context and finding the relation between different entities. \citeauthor{madaan2020eigen} designs an EIGEN model that generates the influence graph for multi-hop reasoning. However, graph models are difficult to deal with the explicit reasoning over qualitative comparison expressions.
In contrast, our RGN model finds the line of reasoning and relations using entity gating module and relation gating modules. These two modules help RGN to reason over qualitative comparisons for describing the changes in the procedural text. We will explain more details about the differences between baseline models and the RGN model in Section~\ref{sec:rgn_model}. 

\begin{figure*}
\centering
\includegraphics[width=0.80\textwidth,height=190pt]{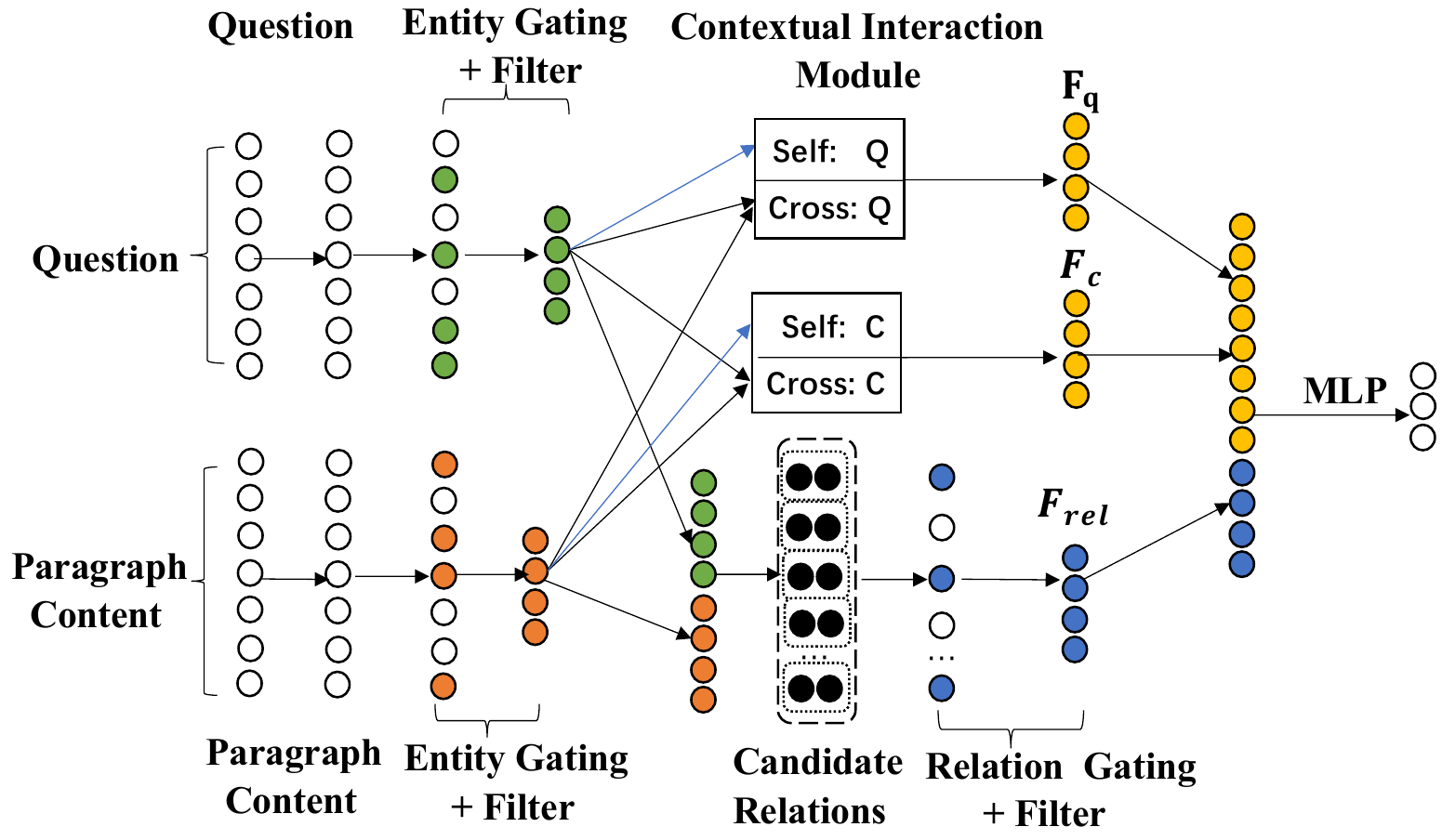}
\caption{Relational Gating Network~(RGN) is composed of pre-training contextual representation, entity gating module,  relation gating module, and contextual interaction module followed by a task-specific classifier.}
\label{fig:architecture}
\end{figure*}

\section{Relational Gating Network}
\label{sec:rgn_model}

Relational Gating Network~(RGN) aims to establish a framework for reasoning over procedural text. The end-to-end network uses an entity gating module to extract and filter the critical entities in question and paragraph content. We enable a higher-order chain of reasoning patterns based on the pairwise relationships between key entities generated from the entity gating and relation gating representations. We propose a practical contextual interaction module to improve the entity alignment in an efficient way.
Figure~\ref{fig:architecture} shows the proposed architecture. This section introduces our network and the training approach in detail. 

\subsection{Problem Formulation}
Formally, the task is to select one of the candidate answers $a$, (A) More; (B) Less; (C) No effect, given a question $q$ and the paragraph content $\mathcal{C}$. The paragraph content includes several sentences $\mathcal{C}=\{s_1, s_2, \dots, s_n\}$. Therefore, for each data sample, the data format is a triplet of ($q$, $\mathcal{C}$, a).

\subsection{Entity Representations}
For each data sample, we form the input $E$ based on the question $q$ and the paragraph content $\mathcal{C}$ as follows: 
\begin{align}
E = [ [CLS];q;[SEP]; \mathcal{C} ],
\end{align}
where [CLS] and [SEP] are the special tokens used in RoBERTa~\cite{Liu2019RoBERTaAR}.
We feed input $E$ to a pre-trained LM to obtain token representations. Then we use $E_{[CLS]}$ representation as the compact representation of the paragraph.
After that, we use $E$ to obtain $E_{[CLS]}$, $E_q$, and $E_\mathcal{C}$, which are shown as follows:

\begin{align}
E_q = [ E_q^{w_1}, E_q^{w_2}, \dots, E_q^{w_m}] \in \mathbb{R}^{m \times d} , \\
E_\mathcal{C} = [ E_\mathcal{C}^{w_1}, E_\mathcal{C}^{w_2}, \dots, E_\mathcal{C}^{w_n} ] \in \mathbb{R}^{n \times d} ,
\end{align}

\noindent where $E_q$ represents the question contextual representation, $E_\mathcal{C}$ represents the paragraph contextual content representation, $d$ is the learned representation dimension for tokens, $m$ represents the max length of the question, and $n$ represents the max length of the paragraph content. Each token will be a candidate \textit{entity}.

\begin{figure*}
\centering
\includegraphics[width=0.80\textwidth,height=190pt]{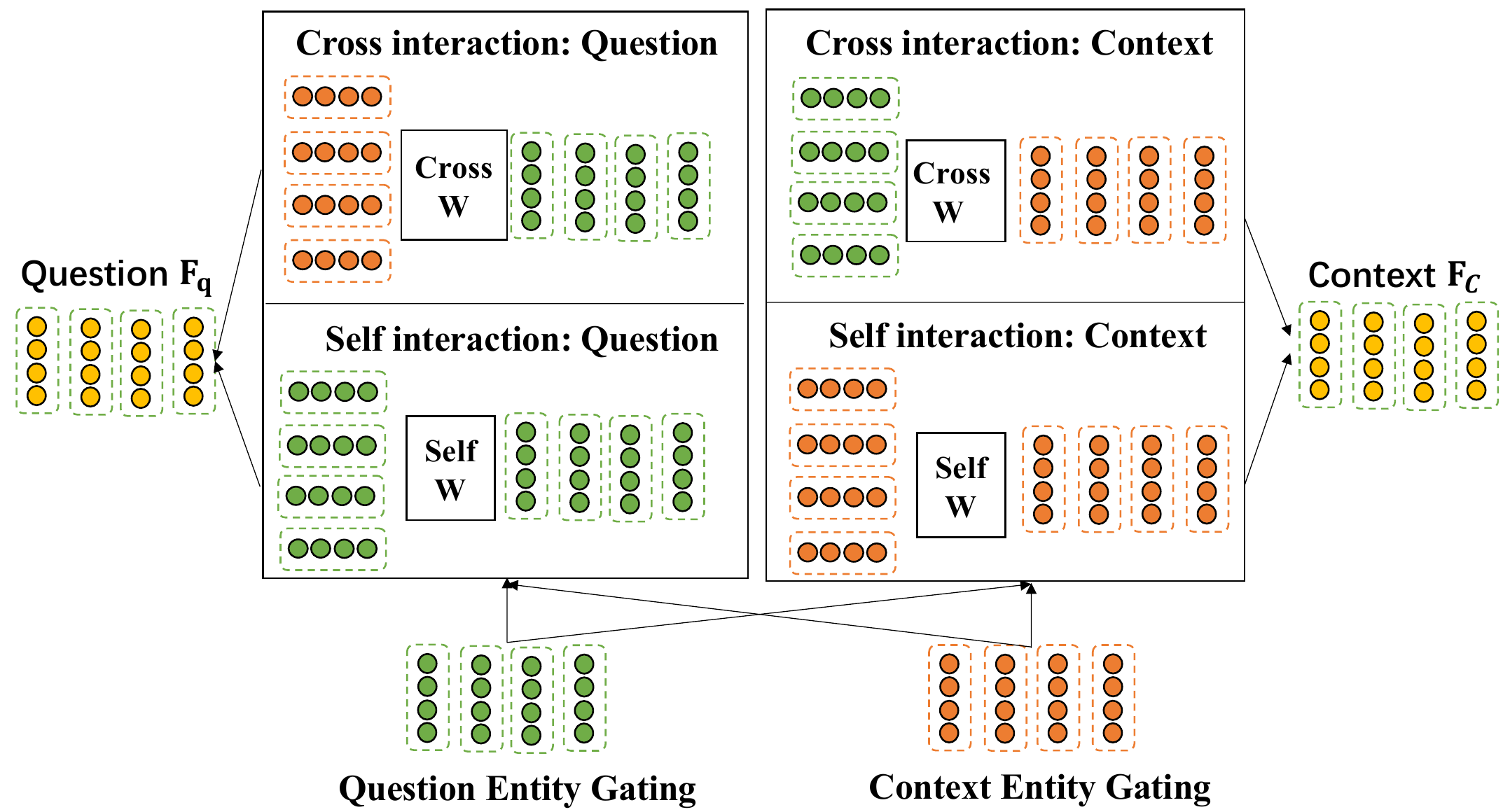}
\caption{Contextual Interaction Module comprises self-interactions and cross-interactions. The inputs are the question's and paragraph's filtered entities representations, and the outputs are question and paragraph contextual representations.}
\label{fig:crm}
\end{figure*}

\subsection{Entity Gating}
\label{sec:entity_gate}

The intuition behind the entity gating module is to filter several key entities representations from both question, $E_q$, and paragraph content, $E_\mathcal{C}$. We call this process entity gating which is shown in Figure~\ref{fig:architecture}. 
Given the question $E_q$, for each entity $E_q^{w_i}$, we use a multi-layer perceptron and a softmax layer to obtain an entity importance score $U_q^{w_i}$:

\begin{align}
U_q^{w_i} &= \frac{ \exp \left( MLP(E_q^{w_i}) \right) }
    {\sum_{j=1}^m \exp \left( MLP(E_q^{w_j})\right) }, \\
E_q' &= U_q E_q \in \mathbb{R}^{m \times d}.
\end{align}

We compute the new entity representations $E_q'$ by multiplying the entity representations and their scores in $U_q$. Then we choose the entities with top-$k$ scores. 
We denote the set of filtered key entities after gating the question as $V_q  = [V_q^{1}, V_q^{2}, \dots, V_q^{k}] \in \mathbb{R}^{k\times d}$.
$V_q$ is the question gated entity representation, $k$ is the number of filtered entities and $V_q^{i}$ is $d$-dimensional embedding for the $i$-th filtered entities.

Notice that the process of computing paragraph entity gating $V_\mathcal{C}  \in \mathbb{R}^{k \times d}$ is the same as the question entity gating $V_q$.
Using the entity gating mechanism improves the interpretability of our deep model as we can explicitly see the selected entities and interpret their qualitative comparisons.
A more detailed analysis is shown in Section~\ref{sec:ablation}.

\subsection{Relation Gating}
``What if'' reasoning requires not only an understanding of the question and paragraph contexts, but also requires reasoning over entities and their relations.
Therefore, we take the representations beyond entities into consideration by using a relation gating module. This extension allows RGN to capture the higher-order chain of reasoning based on pairwise relations, which is one of the main contributions of our paper.
The pairs of entities can help to understand the connections between words and finding the line of reasoning.
Moreover, relation gating aims to pair un-directed relations between entities for capturing the crucial relations, like ``tadpole (losses) tail'' and ``less severe'',
as well as the pairs of entities that help to understand the line of reasoning. 
We call this process relation gating module, which is shown in Figure~\ref{fig:architecture}.

In this module, first, we concatenate $V_q$ and $V_\mathcal{C}$, which are obtained from Section~\ref{sec:entity_gate} and form candidate set $V = \{V_q;V_\mathcal{C} \}$. Then we pair every two gated entities and form $V_{rel}^{i,j}=[V^i; V^j] \in \mathbb{R}^{1 \times 2d}$. 
Furthermore, the candidate relational representation, $V_{rel}$, is a non-linear mapping $\mathbb{R}^{2d} \rightarrow \mathbb{R}^{2d}$ modeled by fully connected layers from candidate relation. 
$$ V_{rel} = [ V_{rel}^1, V_{rel}^2, \dots, V_{rel}^{r} ] \in \mathbb{R}^{r \times 2d},$$  where $r$ is the size of total relation candidate pairs, that is, $r = \frac{2k \times (2k -1 )}{2}$.
Given each candidate relation $V_{rel}^{i}$, we compute a multi-layer perceptron and a softmax layer to obtain a relational importance score, $T^i$:

\begin{align}
T^i &= \frac{ \exp \left( MLP(V_{rel}^i) \right) }
    {\sum_{j=1}^{p} \exp \left( MLP(V^{rel}_j)\right) }, \\
V_{rel}' &= T V_{rel}  \in \mathbb{R}^{k \times 2d}.
\end{align}

We compute the new relation representation $V_{rel}'$ by multiplying the relation representations and their scores in $T$. 
We select the relations with top-$k$ scores. 
Using all the scores increases the number of parameters and the computational cost significantly, moreover, the redundant entities make the learning harder and consequently less accurate.
We denote the set of filtered key relations after gating relations as $F_{rel} \in R^{ k\times 2d} $ to the gated relation representation.

\subsection{Contextual Interaction Module}
Entity alignment is one of the challenges in procedural reasoning. Although we separately propose entity gating and relation gating in the above sections, aligning questions with the paragraph is still important. We found that a simple concatenation of gated entity representations from the question $V_q$ and the paragraph content $V_\mathcal{C}$ shows a good performance. However, concatenated representations and multi-layer perceptrons 
have a limited capacity for modeling the interactions.

As shown in Figure~\ref{fig:crm}, we have developed a novel and fast encoding model, namely Contextual Interaction Module. The model needs to incorporate information from Question-Content interactions, and meanwhile avoid expensive architectures such as Multi-Head attentions~\cite{vaswani2017attention} because those are infeasible for large-scale datasets. Thus, we developed a model that uses only linear projections and inner products of both sides, i.e., question and context, and we apply a mechanism like simplified self-attention to model the interactions as described below.

Given the $V_q$, we compute the self-interaction of the question's gated entities, $F_{q}^{self}$, \[F_{q}^{self} = V_q^T W^{self} V_q \quad \in \mathbb{R}^{k \times d},\]
where $W^{self} \in \mathbb{R}^{d \times k}$ is a projection matrix. 
The cross-interactions between gated entities and paragraph entities can be calculated as
\[F_{q}^{cross} = V_\mathcal{C}^T W^{cross} V_q \quad \in \mathbb{R}^{k \times d},\]
where $W^{cross} \in \mathbb{R}^{d \times k}$ is also a projection matrix. 
Then we concatenate the two matrices $F_{q}^{self}$ and $F_{q}^{cross}$. Finally, we obtain the question contextual representation $F_{q}$ as follows:

\begin{align}
F_{q} = [F_{q}^{self};F_{q}^{cross}] \in \mathbb{R}^{k \times 2d}
\end{align}
Notice that the process of paragraph contextual representation $F_\mathcal{C} \in R^{k \times 2d} $ is the same as the question contextual representation $F_q$.
Therefore, the output includes two representations. One is the paragraph contextual representation containing information from the question, and the question contextual representation containing information from the paragraph. 

\subsection{Output Prediction}

After acquiring all the contextual entity representations and gated relations representations, we concatenate them and use the result as the final representation, $F$. The process is described as below:
\begin{align}
F &= [F_{q}; F_{c}; F_{rel}] \in \mathbb{R}^{3k \times 2d}
\end{align}
Finally, a task-specific classifier $\mbox{MLP}\left( F \right)$ predicts the output. 

\begin{table*}[ht!]
\begin{center}
\begin{tabular}{|l|ccc|c|}
\hline
Models            & in-para  & out-of-para & no-effect & Test V1 Acc \\ 
\hline
 \emph{Majority}      &45.46 &49.47 & 55.0 &30.66 \\
 \emph{Adaboost}~\cite{Devlin2019BERTPO}       & 49.41 & 36.61  &48.42  &43.93   \\
 \emph{Decomp-Attn}~\cite{Parikh2016ADA}      &56.31 &48.56 &73.42 & 59.48\\
\hline
 \emph{BERT (no para)}~\cite{Devlin2019BERTPO}     &60.32 &43.74 &84.18 &62.41 \\
 \emph{BERT}~\cite{Tandon2019WIQAAD}     &79.68 & 56.13 & 89.38 & 73.80 \\
\emph{RoBERTa}~\cite{Tandon2019WIQAAD}     &74.55 &   61.29 & 89.47 & 74.77 \\
 \emph{EIGEN}~\cite{madaan2020eigen}     & 73.58 & 64.04 & 90.84 & 76.92 \\

 \emph{REM-Net}~\cite{huang2020rem}     & 75.67 & 67.98 & 87.65 & 77.56 \\
 \emph{Logic-Guided}~\cite{asai2020logic}     & - & - & - & 78.50 \\
 \textbf{\emph{RGN}} & \textbf{80.32} &\textbf{68.63} &\textbf{91.06} & \textbf{80.18} \\
\hline
  Human     & - & - & - & 96.33\\
\hline
\end{tabular}
\end{center}
\caption{Model Comparisons on WIQA test V1 dataset. WIQA test data has four categories, including in-paragraph accuracy, out-of-paragraph accuracy, no effect accuracy, and overall test accuracy. }
\label{table:results_wiqa_main}
\end{table*}

\subsection{Training Strategy}
We use the cross-entropy loss for training the RGN model.
To update the weights of the Entity Gating module described in Section~\ref{sec:entity_gate}, we use a $0/1$ tensor (1 for top-$k$ indices, 0 for others) to record the top-k indices extracted from the score tensor $U_q$. Next, we use that tensor to multiply the entity representation matrix $E_q'$ to obtain the gated entity representations $V_q$. Therefore, the model can backpropagate the error by updating the learned weights. We use the same method to update the weights of the Relation Gating module. The RGN model can be trained end-to-end with cross-entropy loss.

\section{Experimental Setup}

\subsection{Dataset}

WIQA is a large-scale collection of "what \dots if" reasoning. WIQA dataset\footnote{WIQA dataset is available at \url{http://data.allenai.org/wiqa/}.} contains two parts: paragraphs describing a procedural text and multiple-choice questions. The task is to answer the questions given a paragraph of description.
Table~\ref{table:data_stat} shows the detailed data statistics and data distribution of the WIQA dataset.

\begin{table}[ht!]
\begin{center}
\resizebox{0.47\textwidth}{19mm}{
\begin{tabular}{|ll|cccc|c|}
\hline
Data &       & Train & Dev  & Test V1 & Test V2   & Total \\
\hline
Questions     &      &  29808     &  6894   &  3993 &  3003    &   43698  \\ 
\hline
 & in-para &  7303     &  1655   &    935   &   530  &  10423   \\ 
Question & out-of-para &  12567     & 2941   & 1598  &   1218   &    18326 \\ 
 type        & no-effect & 9936      & 2298    &  1460  &   1255   &    14949 \\
        &   Total    &  29808   &  6894    & 3993   &  3003  & 43698\\
\hline
        & \#hops=0 &9936 &2298 & 1460 & 1255 & 14949   \\ 
Number  & \#hops=1 &6754 &1510 & 835  & 245 & 9254   \\ 
of hops & \#hops=2      &8969 &2145 & 1153 & 1027 & 13294    \\ 
        & \#hops=3     &4149 &941  & 545 & 476 & 6111   \\ 
        &   Total  & 29808   &  6894    & 3993 & 3003  &  43698   \\
\hline
\end{tabular}
}
\end{center}
\caption{WIQA Dataset Statistics.}
\label{table:data_stat}
\end{table}

\begin{table}[ht!]
\begin{center} \small
\resizebox{0.46\textwidth}{20mm}{
\begin{tabular}{|l|ccc|c|}
\hline
 Models            & in  & out & no-eff & Test V2 \\ 
\hline
  \emph{Random}      &33.33& 33.33& 33.33& 33.33 \\
  \emph{Majority}      & 00.00 & 00.00 & 100.0 & 41.80 \\
\hline
  \emph{RoBERTa}     & 70.69 & 60.20 & 91.11 & 75.34\\
    \emph{REM-Net}    & 70.94 & 63.22 & 91.24 & 76.29 \\
  REM-Net (RoBERTa-large) & 76.23 & 69.13 & 92.35 & 80.09 \\
  \emph{QUARTET} & 74.49 & 65.65 & 95.30& 82.07 \\
  \cite{rajagopal-etal-2020-ask}    & & & & \\  
\textbf{\emph{RGN (RoBERTa-base)}} & 75.91 & 66.15 & 92.12 & 79.95 \\
   \textbf{\emph{RGN (RoBERTa-large)}} & \textbf{78.40} & 68.83 & 93.01 & \textbf{82.46} \\
\hline
Human     & - & - & - & 96.30\\
\hline
\end{tabular}
}
\end{center}
\caption{Model Comparisons on WIQA test V2. ``In'' represents in-paragraph accuracy, ``out'' represents out-of-paragraph accuracy, and ``no-eff'' represents no effect accuracy, .}

\label{table:results_wiqa_main_2}
\end{table}

\subsection{Experiment Setting}
We implemented RGN using PyTorch\footnote{Our code is
available at \url{https://github.com/HLR/RGN}.}. 
We used RoBERTa-Base in our model. Therefore, all of the representations are $768-$dimensions.
For each data sample, we keep $128$ tokens as the max length for the question, and $256$ tokens as the max length for paragraph contents. 
Notice that both gated entity representations for question and paragraph use $k=10$ for selecting top-$k$ entities in our experiments. The value of this hyper-parameter was selected after experimenting with various values in $\{3, 5, 7, 10, 15, 20\}$ using the development dataset.
For the Gated relation representations, top-$10$ ranked pairs are used to reduce the computational cost and reduce the unnecessary relations. 
In the relation gating process, we use two hidden layers for multi-layer perceptrons. The task-specific output classifier contains two MLP layers. The model is optimized using the Adam optimizer. The training batch size is $4$. 
During training, we freeze the parameters of RoBERTa in the first two epochs, and we stop the training after no performance improvements observed on the development dataset which happens after $8$ epochs.

\section{Results and Analysis}

We show the model performance on the WIQA task compared to other baselines in Table~\ref{table:results_wiqa_main} and Table~\ref{table:results_wiqa_main_2}. We observe that, in general, Transformer-based models outperform other models, like Deomp-Attn~\cite{Parikh2016ADA}. This promising performance demonstrates the effectiveness of Transformers~\cite{vaswani2017attention} and strong pre-trained contextual representations~\cite{Devlin2019BERTPO,Liu2019RoBERTaAR}. Moreover, our RGN achieves state-of-the-art results compared to all other models. Especially, RGN outperforms~\citeauthor{Tandon2019WIQAAD} by $6.38\%$ and outperforms current state-of-the-art model on test V1, logic-guided~\cite{asai2020logic}, by around $1.6\%$. Moreover, our RGN model achieves the SOTA on WIQA test V2. The improved performance demonstrates that entity gating, relation gating, and contextual interaction module are effective for ``what...if'' procedural reasoning. We provide a detailed analysis of the advantage of RGN from different perspectives. 


\begin{table}[!ht]
\begin{center}
\resizebox{0.45\textwidth}{11mm}{
\begin{tabular}{|l|c|c|c|}
\hline
Model  & \# hops = 1  &  \# hops = 2 & \# hops = 3 \\ 
\hline
BERT(no para) & 58.1\% & 47.3\% & 42.8\% \\
BERT & 71.6 \% & 62.5\% & 59.5\% \\
RoBERTa & 73.5 \% & 63.9\% & 61.1\% \\
EIGEN    & 78.78  \% & 63.49\% & 68.28 \% \\
RGN & \textbf{80.5\%} & \textbf{71.2\%} & \textbf{70.0\%} \\
\hline
\end{tabular}
}
\end{center}
\caption{The accuracy when the number of hops increases.}
\label{table:num_of_hops}
\end{table}

\begin{figure*}
\centering
\includegraphics[width=0.95\textwidth,height=170pt]{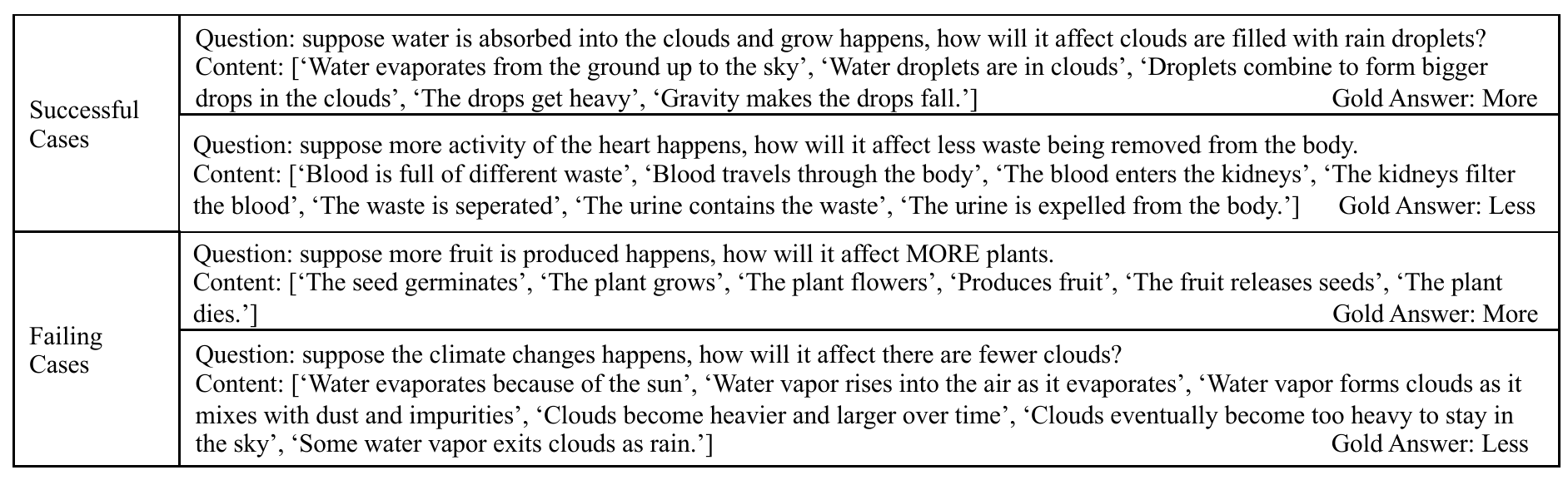}
\caption{Successful and failing cases of RGN network.}
\label{fig:case_study}
\end{figure*}

\noindent \textbf{Effects on Relational Reasoning and Multi-Hops}: In-para and out-of-para question categories require multiple hops of reasoning to answer the questions.
As shown in Table~\ref{table:num_of_hops}, we found that the accuracy improved $7.0\%$ for $1$ hop, $7.3\%$ for $2$ hops, and $8.9\%$ for $3$ hops compared to RoBERTa which does not have the two gating mechanisms and Contextual Interaction Module. 
As we expect, the RGN framework has made a huge progress in reasoning with multiple hops and the improvement in the performance of the baselines is more when the number of hops increases.
For qualitative analysis, we show successful cases from our RGN in Figure~\ref{fig:case_study}. We observe that RGN is capable of bridging question and paragraph content by extracting key entities. In the successful cases, which is shown in Figure~\ref{fig:case_study}, RGN helps in constructing the chain of ``water droplets are in clouds $\rightarrow$ droplets combine to form bigger drops in the clouds'' through key entities ``water'', ``clouds'', and ``droplets''. Moreover, we observe that the key entities  ``water'', ``clouds'', and ``droplets'' also obtain high attention scores.

\noindent \textbf{Effects on Qualitative Comparisons}: Like \citeauthor{Tandon2019WIQAAD} said, around $65\%$ of the change statements make use of qualitative comparisons to describe the changes in the procedure. 
For example, change statements use comparative words such as (more, less), (colder, hotter).
Besides, the change statement for a positive or negative influence is context-dependent, making the ``what \dots if'' reasoning challenging.
The qualitative comparisons are improved in both in-para, out-of-para, and no-effect using RGN. The results are shown in Table~\ref{table:results_wiqa_main_2}. We observe that RGN outperforms $6.74\%$ on in-para and $4.59\%$ on out-of-para compared to EIGEN~\cite{madaan2020eigen} -a very recent baseline.


\subsection{Ablation Study}
\label{sec:ablation}

\noindent \textbf{Effects of Entity Gating:} In the first study, we remove the entity gating and relation gating modules. 
Notice that the contextual interaction module uses the whole question entities and paragraph entities when RGN does not use these two modules. 
Using whole entities significantly increases the computational cost.
Moreover, Table~\ref{table:ablation} shows that the accuracy is lower about $5.3\%$ compared to full RGN when applied on the development dataset. This experiment demonstrates that using all the entities without gating mechanism has a negative influence on the contextual interaction module and drops the performance.

\begin{table}[!ht]
\begin{center}\small
\resizebox{0.46\textwidth}{14mm}{
\begin{tabular}{|c|ccc|c|}
\hline
Ablations & in  & out & no-eff  & dev acc \\
\hline
RGN~(w/o gating ent \& rel)  & 76.2 & 61.1 & 89.2 & 75.3  \\
RGN~(w/o gating rel) & 78.4 & 63.6 & 89.9 & 77.4  \\
RGN &  81.7 & 69.2 & 91.3 & 80.6 \\
\hline
RGN~(w/o CIM) & 80.2 & 68.4 & 90.5 & 79.7 \\
RGN~(- CIM + Multi-Head) & 81.3 & 68.9 & 91.7 & 80.3 \\
RGN~(add regularization) & 82.0 & 69.1 & 91.6 & 80.8  \\
\hline
\end{tabular}
}
\end{center}
\caption{Ablation Study. CIM: Contextual Interaction Module. 
}
\label{table:ablation}
\end{table}

\noindent \textbf{Effect of Relation Gating:} 
The goal of the relation gating is to capture the higher-order chain of reasoning based on pairwise relations. 
The relation gating module extracts the important candidate relations by pairing up entities after gating entities.
More importantly, the pairs of entities can help understanding the connections between words and finding the line of procedural reasoning.
Our model captures the important pairs of influencing entities ``tadpole (losses) tail'' and ``animal (hunts) frog''.

When we keep the entity gating module and remove the relation gating module, we observe that the accuracy of WIQA decreases $3.3\%$ compared to the full RGN framework. 
Moreover, the model without the relation gating module can not capture the key relations.
The results show that thee performance on the out-of-para questions decreases $5.6\%$ compared to the full RGN model. Section~\ref{sec:qualitative_ana} shows more examples and analysis.

\noindent \textbf{Effects of Contextual Interaction Module (CIM):}
\citeauthor{Tandon2019WIQAAD} shows that around $15\%$ of the influence changes have difficulties for handling the entity alignment part due to the language variability. In other words, paragraph entities use different terms, such as (“removes”, “expels”) for expressing the same semantics. 
Especially, the problem of language variability becomes more severe for the multi-hop cases that require aligning the question with several sentences in the paragraph. Without the Contextual Interaction Module, the development accuracy decreases more than $1\%$.
As shown in Table~\ref{table:num_of_hops}, the accuracy improves significantly in the direct effect ($1$ hop) and indirect effects ($2$ hops or $3$ hops) compared to all strong baselines. This demonstrates the effectiveness of the interaction module.
In an additional experiment, we replaced the CIM with the Multi-Head attention that uses an encoder of the Multi-Head attention composed of a stack of N = 6 identical layers. Each layer has two sub-layers. The first layer is multi-head self-attention, and the second is a fully connected network~\cite{vaswani2017attention}. The computational time was 936 (ms/batch) for our contextual interaction module while it is 3002 (ms/batch) for the Transformer while the accuracy is fairly similar.

\noindent \textbf{Effects of Regularization:}
To improve the capacity of relational reasoning challenge described in section~\ref{sec:intro}, we follow ~\citeauthor{asai2020logic} work that augments the data by making variations in the questions and imposing consistency between answers of those questions. It uses a regularization term in the learning objective (based on cross entropy loss denoted as $L_{original}$) and the regularization term denoted as $L_{consistency}$), called inconsistency loss. Therefore, the whole loss function is shown as follows:
$$ L = L_{original}(X) + L_{consistency}(X). $$
We found that the RGN model with consistency regularization has about $0.1\% - 0.15\%$ further improvement. 

\subsection{Qualitative Analysis}
\label{sec:qualitative_ana}
For a better understanding of how our proposed model performs qualitatively, we show successful cases and failing cases from our RGN framework in Figure~\ref{fig:case_study}. We can observe that RGN is surprisingly capable of bridging the question and content in the in-para category.

Although the RGN framework has achieved state-of-the-art performance, the framework cannot always capture the line of reasoning. The bottom part of Figure~\ref{fig:case_study} shows some failing cases. In the first failing case, RGN gives a wrong prediction because the content sentence ``the plant dies'' is captured as a strong negative influence by our model. Although our model bridges the relation between ``fruit'' and ``plant'', the critical term ``dies'' obtains a high gating score and misleads our final prediction.

Commonsense reasoning is the other type of error made by RGN. Out-of-paragraph questions require commonsense external knowledge. For example, in the second failing case of Figure~\ref{fig:case_study}, the question contains ``climate change'' and the paragraph does not contain the cause of the ``climate change''. This needs external knowledge between ``climate change'' and ``water evaporate''.
Since answering the question requires external knowledge, it is hard to build a casual relationship for this example.
However, the improvement on the out-of-paragraph is due to observing multiple examples in the dataset that use the same type of commonsense. Because the relational gating helps to find the line of reasoning, our RGN model captures those from observing the relationships frequently and learns shortcuts. For example, in the second successful case of figure~\ref{fig:case_study}, the relational gating module captures the pairwise relation between ``heart body'' and ``blood body'' due to multiple occurrences in the data  --filling the information gap for reasoning.

\section{Conclusion}
Our paper proposes an end-to-end Relational Gating Network~(RGN) to help ``what \dots if'' procedural reasoning over text for answering cause and effect questions. 
Particularly, we propose an entity gating module, relation gating module, and contextual interaction module to find the answer.
We demonstrate that the proposed approach can effectively solve the challenges in the ``what \dots if'' reasoning, including multiple-hop reasoning, qualitative comparison, and entity alignment. 
We successfully evaluate our RGN on the WIQA dataset and achieve state-of-the-art performance. Our gating mechanism and contextual interaction module can be easily used in solving various QA tasks that need to reason over entities and their relationships and follow a procedure.
The gating mechanism can be extended to work at various levels of granularity such as sentence and paragraph levels to filter important pieces of information and to find the line of reasoning for answering the questions.

\section*{Acknowledgments}
This project is supported by National Science Foundation (NSF) CAREER award $\#$2028626 and partially supported by the Office of Naval Research (ONR) grant  $\#$N00014-20-1-2005. Any opinions, findings, and conclusions or recommendations expressed in this material are those of the authors and do not necessarily reflect the views of the National Science Foundation nor the Office of Naval Research.
We thank all reviewers for their thoughtful comments and suggestions.

\bibliographystyle{named}
\bibliography{ijcai21}

\end{document}